\documentclass[review]{elsarticle}

\usepackage{siunitx}
\usepackage{todonotes}
\usepackage{hyperref}
\graphicspath{{./figs/}}

\journal{XXX}
\biboptions{authoryear}

\title{Characterizing the dynamics of learning \\in repeated reference games}

\author[1]{Robert D. Hawkins\corref{mycorrespondingauthor}}
\author[1]{Michael C. Frank}
\author[1,2]{Noah D. Goodman}

\address[1]{Department of Psychology, Stanford University}
\address[2]{Department of Computer Science, Stanford University}

\cortext[mycorrespondingauthor]{Corresponding author. Robert Hawkins, Department of Psychology, Princeton University, Princeton, NJ 08540, \texttt{robertdh@princeton.edu}}

\begin{document}

\begin{frontmatter}





\begin{abstract}
The language we use over the course of conversation changes as we establish common ground and learn what our partner finds meaningful.
Here we draw upon recent advances in natural language processing to provide a finer-grained characterization of the dynamics of this learning process.
We release an open corpus ($>$15,000 utterances) of extended dyadic interactions in a classic repeated reference game task where pairs of participants had to coordinate on how to refer to initially difficult-to-describe tangram stimuli.
We find that different pairs discover a wide variety of idiosyncratic but efficient and stable solutions to the problem of reference.
Furthermore, these conventions are shaped by the communicative context: words that are more discriminative in the initial context (i.e. that are used for one target more than others) are more likely to persist through the final repetition.
Finally, we find systematic structure in how a speaker's referring expressions become more efficient over time: syntactic units drop out in clusters following positive feedback from the listener, eventually leaving short labels containing open-class parts of speech.
These findings provide a higher resolution look at the quantitative dynamics of \emph{ad hoc} convention formation and support further development of computational models of learning in communication. 

\end{abstract}

\end{frontmatter}
\section{Introduction}\label{introduction}

Human language use is remarkably flexible.
We are able to coax new meanings out of existing words --- or even coin new ones --- to handle the diverse challenges encountered in everyday communication \citep{Clark83_NonceSense,davidson_nice_1986}.
This flexibility is partially explained by \emph{de novo} pragmatic reasoning, which allows listeners to use context to infer an intended meaning even in cases of ambiguous or non-literal usage \citep{LascaridesCopestake98_PragmaticsWordMeaning,Glucksberg01_FigurativeLanguage,GoodmanFrank16_RSATiCS}.
However, a rich theoretical thread has suggested that learning mechanisms may also play an important role, allowing speakers and listeners to dynamically adapt their representations of meaning over the course of an interaction \citep{BrennanClark96_ConceptualPactsConversation,pickering2004toward,delaney2019neural}.

Two functional considerations motivate the need for continued learning in communication, even among adults.
First, just as there is substantial phonetic variability across speakers with different accents \citep{kleinschmidt2019structure}, words may vary in meaning from speaker to speaker.
This variability is clear for cases like slang, technical lingo, nicknames, or colloquialisms \citep[e.g.][]{Clark98_CommunalLexicons}, but may extend even to more ordinary nouns and adjectives.
It may be difficult to know at the outset of an interaction exactly which meanings will be shared and which will not, requiring ongoing adaptation.
Second, because we live in a changing environment, we often experience novel entities, events, thoughts, and feelings that we want to talk about but do not already share (literal) words to express.
Both of these obstacles can be overcome using feedback from one's partner to dynamically re-calibrate expectations about meaning.

The \emph{repeated reference game} task has provided a natural and productive paradigm for eliciting behavior under such conditions.
In this task, pairs of participants are presented with arrays of novel images.
On each trial, one player (the director) is privately shown a \emph{target} object and must produce a referring expression allowing their partner (the matcher) to correctly select that object from the array.
The director is then given feedback at the end of each trial about which object the matcher selected, and the matcher is given feedback about the true target object.
Critically, each object appears as the target multiple times in the trial sequence, allowing the experimenter to examine how referring expressions change as the director and matcher accumulate shared experience.
To the extent that the director and matcher converge on an accurate system of stable referring expressions, and these referring expressions differ from the ones that were initially produced, it may be claimed that \emph{ad hoc conventions} or \emph{pacts} have formed within the dyad \citep{hawkins2018emergence}.

One of the earliest and most intriguing phenomena observed in this task is that descriptions are dramatically shortened across repetitions: an initial description like ``the one that looks like an upside-down martini glass in a wire stand'' may gradually converge to ``martini'' by the end  \citep{KraussWeinheimer64_ReferencePhrases}.
That is, speakers are able to communicate the same referential content much more efficiently over time.
Subsequent work has established a number of signature properties of this process through careful experimental manipulation.
First, the extent to which descriptions are shortened is contingent on evidence of understanding from the matcher \citep{KraussWeinheimer66_Tangrams,KraussEtAl77_AudioVisualBackChannel,HupetChantraine92_CollaborationOrRepitition}, and is therefore not easily explained as a mere practice or repetition effect.
Second, the resulting labels are \emph{partner-specific} in the sense that they do not transfer if a novel matcher is introduced \citep{WilkesGibbsClark92_CoordinatingBeliefs,MetzingBrennan03_PartnerSpecificPacts,brennan_partner-specific_2009}.
Third, they are \emph{sticky} in the sense that they persist through precedent with the same partner even after the referential context changes \citep{BrennanClark96_ConceptualPactsConversation}, and are readily extended to similar objects \citep{MarkmanMakin98_ReferentialCommunicationCategory}.
These qualitative effects provide an empirical backbone for theories of communication to explain. 
However, as theories are increasingly formalized as computational models making more precise quantitative predictions, setting criteria to distinguish between them will depend critically upon resolving more detailed theoretical questions about the dynamics of adaptation in natural language communication.

In this paper, rather than arguing in favor of particular theory, we release a new, open corpus of repeated reference games and conduct a variety of analyses to address current gaps in measurement and establish a firmer theoretical foundation facilitating future modeling work.
In particular, we address two methodological challenges that have limited the ability of previous studies to provide a sufficiently fine-grained characterization of behavior.
First, we need more data.
Recent technical developments have allowed interactive multi-player experiments to be run on the web \citep{Hawkins15_RealTimeWebExperiments}, boosting sample sizes by an order of magnitude.
For comparison, seminal work by \cite{ClarkWilkesGibbs86_ReferringCollaborative} used a sample of 8 pairs of participants, while our confirmatory sample alone contains 83 pairs.
Second, the computational techniques needed to work with rich natural language data were limited at the time of prior work, but have become newly tractable given developments in natural language processing (NLP).

Our analyses roughly divide into two broad categories, corresponding to the dynamics of syntactic \emph{structure} and semantic \emph{content}.
Our investigations of syntactic structure in Section \ref{sec:structure} focus on the process by which referring expressions are shortened to communicate the same idea more efficiently.
One particularly simple model, for example, might predict that shortening is purely driven by a random corruption process: at each repetition, each word from the previous repetition's utterance has some probability of being dropped.
Raw word counts alone are not sufficient for disambiguating this simple model from more cognitively complex proposals.
To move beyond word counts, we extracted part-of-speech tags and syntax trees from the text to understand which parts of utterances were being dropped, and in which sequence.
In contrast to the predictions of the random corruption model, we find that clauses and modifiers tend to be dropped in clusters, preferentially leaving open-class parts of speech (e.g. an adjective and noun) by the final repetition, and that the choice to shorten an utterance or not depends on sources of listener feedback.  

In Section \ref{sec:content}, we examine the semantic content of utterances over the course of this shortening process.
Our revolve around the theoretical constructs of \emph{arbitrariness} and \emph{stability}, which have been central to accounts of convention since \cite{Lewis69_Convention}.
Arbitrariness refers to the claim that multiple equally successful solutions exist in the space of possible conventions: there is no single optimal solution that all speakers should objectively use.
Stability refers to the claim that, once a solution has been found, speakers should not deviate from it.
Our contribution is to operationalize these claims in the high-dimensional space of vector embeddings for referring expressions (i.e. GloVe embeddings).
By measuring the similarity between referring expressions in this space, we find that signatures of arbitrariness and stability gradually increase over the course of the interaction.
We also clarify the (non-arbitrary) processes shaping which words eventually become conventions.
In particular, we test the prediction that pragmatic pressures to be informative in context lead more discriminative words to conventionalize \citep{KirbyTamarizCornishSmith15_CompressionCommunication,GibsonEtAl17_ColorNamingUse}.
Taken together, our findings characterize core processes operating within the microcosm of dyadic, natural-language interactions. These processes may ultimately contribute to the adaptive properties of conventions shared across a language community.



\section{Methods: Repeated reference experiment}

We developed two variants of the repeated reference task used in classic work by \cite{ClarkWilkesGibbs86_ReferringCollaborative}: a relatively unconstrained \emph{free-matching} version that more closely replicates classic in-lab designs, and a more tightly controlled \emph{cued} version.
Most importantly, the \emph{cued} version allows us to identify which object each utterance refers to, supporting higher-resolution analyses at the object-by-object level.
We considered the free-matching version to be an exploratory pilot sample and subsequently pre-registered planned analyses for the \emph{cued} version at \mbox{\texttt{\url{https://osf.io/vzvmf}}}.
While we are releasing the corpora from both versions, we restrict our analyses to the \emph{cued} version throughout the paper as a cleaner confirmatory sample.
Any exploratory analyses and changes to our pre-registered plan are described as relevant in each section below.

\begin{figure}
\centering
\vspace{-1em}
\includegraphics[scale=.87]{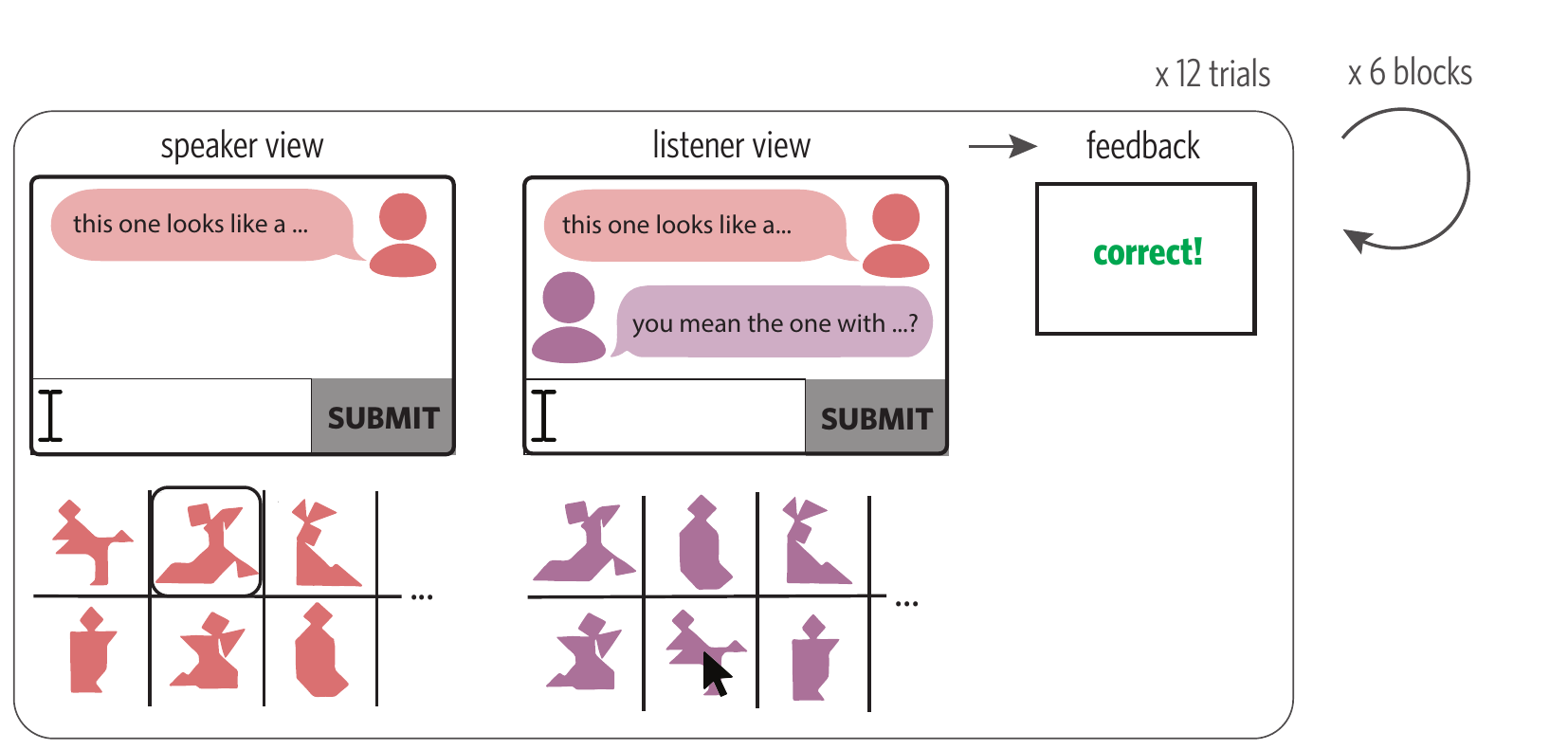}
\caption{Display and procedure for the repeated reference game task.}
\label{fig:design}
\end{figure}

\subsection{Participants}\label{participants}

A total of 480 participants (218 in the \emph{free-matching} version and 262 in the \emph{cued} version) were recruited from Amazon's Mechanical Turk and paired into dyads to play a real-time communication game.

\subsection{Stimuli \& Procedure}\label{stimuli}

On every trial, participants were shown a \(6 \times 2\) grid containing twelve tangram shapes (see Fig. \ref{fig:design}), reproduced from \cite{ClarkWilkesGibbs86_ReferringCollaborative}.
After passing a short quiz about task instructions, participants were randomly assigned the role of either `director' or `matcher' and automatically paired into virtual rooms containing a chat box and the grid of stimuli.
The chat box was a standard modern messaging interface. 
Participants saw a binary indicator that their partner was currently typing, but only saw the message once their partner hit the \emph{Enter} key or clicked the \emph{Send} button.
Use of the chat box was completely unrestricted in both versions of the task: both participants could freely use the chat box to communicate at any time and there was no limit on the number or length of messages that could be sent.

In the \emph{free-matching} version, our procedure closely followed \cite{ClarkWilkesGibbs86_ReferringCollaborative}.
The director and matcher began each trial with scrambled boards.
The director's tangrams were fixed in place, but the matcher's could be clicked and dragged into new positions.
The players were instructed to communicate through the chat box such that the matcher could rearrange their shapes to match the order of the director's board.
When the players were satisfied that their boards matched, the matcher clicked a `submit' button that gave players batched feedback on their score (out of 12) and scrambled the tangrams for the next round.
After six rounds, players were redirected to a short exit survey.
Cells were labeled with fixed numbers from one to twelve in order to help participants easily refer to locations in the grid.

While this replicated design allowed highly naturalistic interaction, it posed several problems for text-based analyses.
First, utterances must contain not only descriptions of the tangrams but also information about the intended location (e.g. '\emph{number 10} is the \dots').
Additionally, because there were no constraints on the sequence, participants could revisit tangrams out of order or mention multiple tangrams in a single message, making it difficult to isolate exactly which utterances referred to which tangrams without extensive hand-annotation.
Finally, the design of the `submit' button made it easy for players to occasionally advance to the next round without referring to all 12 tangrams.

To address these problems, we designed a more straightforwardly sequential \emph{cued} variant of the task design where directors were privately cued to refer to targets one-by-one and feedback was given on each trial (Fig. \ref{fig:design}).
This additional structure allowed us to conduct analyses at the object-by-object level.
On each trial, one of the twelve tangrams was privately highlighted for the director as the \emph{target}.
Instead of clicking and dragging into place, matchers simply clicked the one they believed was the target.
They were not allowed to click until after a message was sent by the director, but were not restricted in their use of the chat box.
We constructed a sequence of six blocks of twelve trials (for a total of 72 trials), where each tangram appeared once per block.
Because targets were cued one at a time, numbers labeling each square in the grid were irrelevant and we removed them.
The grid of tangrams was scrambled on every trial, and participants were given full, immediate feedback: the director saw which tangram their partner clicked, and the matcher saw the intended tangram.

\subsection{Exclusion criteria}

After applying our pre-registered exclusion criteria for games that terminated before the completion of the experiment due to server error or network disconnection (40 in \emph{free matching} and 33 in \emph{cued}), as well as games where participants reported a native language different from English (2 in \emph{free matching} and 3 in \emph{cued}), we implemented an additional exclusion criterion based on accuracy.
We used a 66/66 rule, excluding pairs that got fewer than 66\% of trials correct ($\le8$ of 12)  on more than 66\% of blocks ($\ge4$ of 6).
While most pairs were near ceiling accuracy by the final repetition, this rule excluded 11 in \emph{free matching} and 8 in \emph{cued} who appeared to be guessing or rushing to completion.

\subsection{Data pre-processing}

We used a three step pre-processing pipeline to prepare this corpus for subsequent analyses. Unless otherwise noted, we used the open-source Python package \texttt{spaCy} (version 2.2) to implement all NLP analyses.

\begin{enumerate}

\item \textbf{Spell-checking and regularization}: We conservatively extracted all tokens that did not exist in the vocabulary of the smallest available ($\sim$ 50,000 word) \texttt{spaCy} model and passed them through the SymSpell spell-checker.\footnote{Available at \texttt{https://github.com/wolfgarbe/SymSpell}. We used the smallest model because larger models include typo forms (e.g. `teh') in their vocabulary and thus are not useful for catching errors.} These suggested corrections were then sequentially presented to the first author and either accepted or overridden at their judgement. This process constructed a spell-correction dictionary containing 677 corrections.

\item \textbf{Cleaning unrelated discourse}: Because we allowed our participants to interact in real-time through the chat box, many pairs produced text unrelated to the task of referring to the current target (e.g. greeting one another, asking personal questions, commenting on the length of the task or the results of previous trials). We wanted to ensure that our results were not confounded by patterns in this kind of discourse across the task, and that the semantic content we observe on a particular trial is in fact being used to refer to the current target rather than task-irrelevant topics or, as we found in some cases, referring to other tangrams while debriefing previous errors. We therefore conducted a manual review removing any text not directly referring to the current target. For example, utterances like ``the dancing woman'' and ``this is the one we got wrong last time'' were kept in because they were referring to properties of the current tangram, but words like ``yeah'' or ``ok'' and messages like ``good job'' and ``they'll go quicker if you remember what I say!'' were not. This review affected 1,448 messages, and we also saved these corrections in a dictionary.

\item \textbf{Collapsing multiple messages within a trial}: Finally, some directors used our chat box like an texting interface, hitting the enter key between every micro-phrase of text. Because many of our analysis techniques depend upon having a single referring expression for each trial, we collapsed all messages sent by each participant within a trial into a single message by inserting commas between messages. We chose to use commas because it tends to maintain grammaticality and does not inflate word counts. Importantly, we used this strategy even on trials where there were multiple rounds of interaction between the director and matcher, joining together a director's original message with their follow-up to the matcher's response. This strategy had the advantage of preserving all of the referential content produced by the director on a trial, compared to a `purer' strategy of removing messages sent in response to the matcher, but the disadvantage of losing the precise interactional structure of the dialogue. 

\end{enumerate}

\noindent After implementing exclusions and cleaning, we were left with a \emph{free matching} corpus containing a total of 8,639 ($\approx$ 50,000 words) messages over 56 complete games and a \emph{cued} corpus containing 7,867 messages ($\approx$ 46,000 words) over 83 games.


\section{Results: characterizing the dynamics of structure}
\label{sec:structure}

\begin{figure}[t]
\centering
\includegraphics[scale=.55]{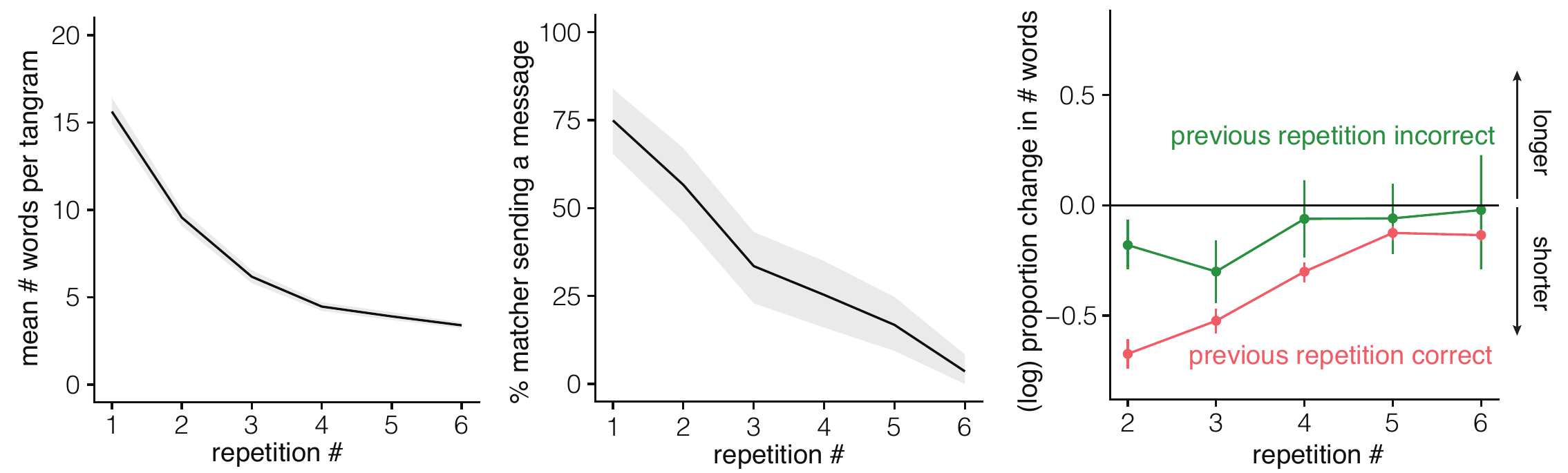}
\caption{(A) Directors use fewer words per tangram over time, (B) matchers are less likely to send messages over time, and (C) directors are sensitive to feedback from the matcher's selection, modulating the reduction in message length on the subsequent repetition of a tangram after an error is made.}
\label{fig:feedback}
\end{figure}

Our first set of analyses examines how the structure of participants' utterances changes over the course of our experiment.
We begin with the observation that the mean number of words used by directors for each tangram decreases strongly over time (see Fig. \ref{fig:feedback}A).\footnote{A similar reduction curve was found in the  ``free matching'' version of the task, though it required more words overall. Participants needed to additionally mention the spatial location where the tangram needed to be moved (i.e. ``number 3 is the \dots``).}
This result replicates a highly reliable reduction effect found throughout the literature on repeated reference games \citep[e.g.][]{KraussWeinheimer64_ReferencePhrases,BrennanClark96_ConceptualPactsConversation}, though participants in our task used fewer words overall than reported by \cite{ClarkWilkesGibbs86_ReferringCollaborative}.
This difference is likely due to the text-based (vs.~spoken) interface.
The following analyses break down this general gain in efficiency into a finer-grained set of phenomena concerning the \emph{structure} of referring expressions over time.
What sequence of transformations do descriptions undergo over the course of repeated reference?

\subsection{The effect of listener feedback on reduction}\label{listener-feedback}

Conventions are formed \emph{collaboratively}, not in isolation \citep{ClarkWilkesGibbs86_ReferringCollaborative,SchoberClark89_Overhearers}, and thus depend on some form of social feedback.
We consider two channels of feedback that were available to speakers.
First, matchers could voluntarily initiate a bi-directional feedback process at any point within a trial by asking follow-up questions, suggesting corrections, and acknowledging or verbally confirming their own understanding through a backchannel.
Second, we automatically supplied ground-truth feedback about the matcher's selection and the true target at the end of each trial.
We expect that speakers integrate both sources of evidence about the listener's understanding to determine how to shorten their utterances.
If either feedback channel is restricted, or if they provide conflicting evidence, descriptions may not reduce as strongly \citep{KraussWeinheimer66_Tangrams, HupetChantraine92_CollaborationOrRepitition, GarrodFayLeeOberlanderMacLeod07_GraphicalSymbolSystems}.

We predicted that the matcher's use of backchannel feedback should be highest on the first repetition and drop off once meanings are agreed upon, consistent with the patterns observed by \citep{ClarkWilkesGibbs86_ReferringCollaborative}.
To test this prediction, we coded whether the matcher sent a message or not on each trial and fit a mixed-effects logistic regression model with a fixed effect of repetition, random intercepts and slopes for each pair of participants, and a random intercept for each target.
We found that the probability of the matcher sending a message decreased significantly over the game $(b=-0.84, t = -9.1, p < 0.001$).
While usage of the backchannel in our online text-based task was less frequent overall than reported in previous verbal lab experiments, we nonetheless strongly replicated the overall trend.
In aggregate, 75\% of matchers responded with at least one message over the twelve trials in the first repetition block, but only 4\% sent a message in the last block (see Fig. \ref{fig:feedback}B).
These messages were frequently questions: as a lower bound, we observed that 49\% of matcher messages explicitly contained question marks (e.g. ``is it standing?'')
Other messages simply echoed the director's label or suggested alternative labels\footnote{We also pre-registered an analysis examining whether a higher rate of listener messages on early rounds would lead to greater overall reduction in speaker descriptions, but later realized that this analysis was confounded. When listeners send more messages on early rounds, the speaker produced more words in response, which led by definition to greater reduction from this higher initial verbosity.}.

Next, as an exploratory analysis, we examined the extent to which directors were sensitive to the ground-truth feedback that was provided at the end of each trial about which tangram the matcher actually \emph{selected}.
If the matcher failed to select the correct target, the director may take this as evidence that their description was insufficient and attempt to provide more detail the next time they must refer to the same tangram.
If the matcher is correct, on the other hand, the director may take this as evidence of understanding and reduce their level of detail when the tangram next appears.
Note that ground-truth feedback provided the speaker distinct information from backchannel feedback within the trial: backchannel feedback did not guarantee a correct response, and matchers often made the correct response without sending any messages in response. 
For example, errors on the first repetition block were only slightly less likely when matchers engaged in dialogue through the chatbox (20\%) than when they stayed silent (23\%; difference not significant, $\chi^2(1) = 0.56, p = 0.45$), although matchers may also have been more likely to initiate backchannel responses on more difficult trials.

We tested the speaker's sensitivity to ground-truth feedback by comparing the proportional change in utterance length (i.e. $\log(n_{t}/n_{t-1})$) on the block after an error against the change after a correct response.
This measure could be positive, indicating a net increase in utterance length, or negative, indicating a reduction.
We fit a mixed-effects regression model predicting this measure with a categorical fixed effect of the matcher's response for the same at the previous repetition block (correct vs. incorrect) and a (centered) continuous effect of repetition block number, including maximum random effects at the speaker level.
We found a significant main effect of feedback, controlling for block number: utterance length decreased more after correct responses than after incorrect responses, $b = -0.28, t = -7.2, p < 0.001$ (see Fig. \ref{fig:feedback}C).

Although appearances of the same tangram were spaced out by block, it is still possible that this effect is not item-specific but the result of lower level attentional or affective mechanisms triggered in the aftermath of an error signal.
To evaluate this possibility, we also measured the proportional change in utterance length on the following \emph{trial}, when feedback about the matcher's response would be freshest but the target tangram would be different.
We then constructed a second regression model including categorical fixed effects of matcher response (correct vs. incorrect) and item-specificity (change measured relative to previous trial vs. previous repetition block), as well as their interaction, with no random effects.
We found a significant cross-over interaction, $b = -0.32, t = -6.2, p <0.001$.\footnote{We report the results of a traditional linear regression model because even the most minimal random effect structure encountered singularity issues during optimization. Because matcher errors were relatively infrequent, these singularities were likely caused by an asymmetry in cell size between the correct and incorrect levels of the matcher response variable. However, when we fit a Bayesian regression with maximal random effects, using the default priors implemented by the \texttt{brms} package to prevent variances from collapsing to boundary values, we found a nearly identical estimate of the interaction coefficient, $b = -0.31$, 95\% credible interval: $[-0.42, -0.19]$.}
The sensitivity to feedback we observed on the subsequent repetition block is not present on the subsequent trial: speakers are equally likely to use more or less words immediately after a correct response, and actually use slightly fewer words on the trial immediately after an \emph{incorrect} response due to a regression to the mean: statistically, more words than average are used for harder tangrams.
This pattern of results is consistent with sensitivity to \emph{tangram-specific} evidence of the matcher's understanding when deciding to modify referring expressions.


\begin{figure}[t!]
\centering
\includegraphics[scale=.8]{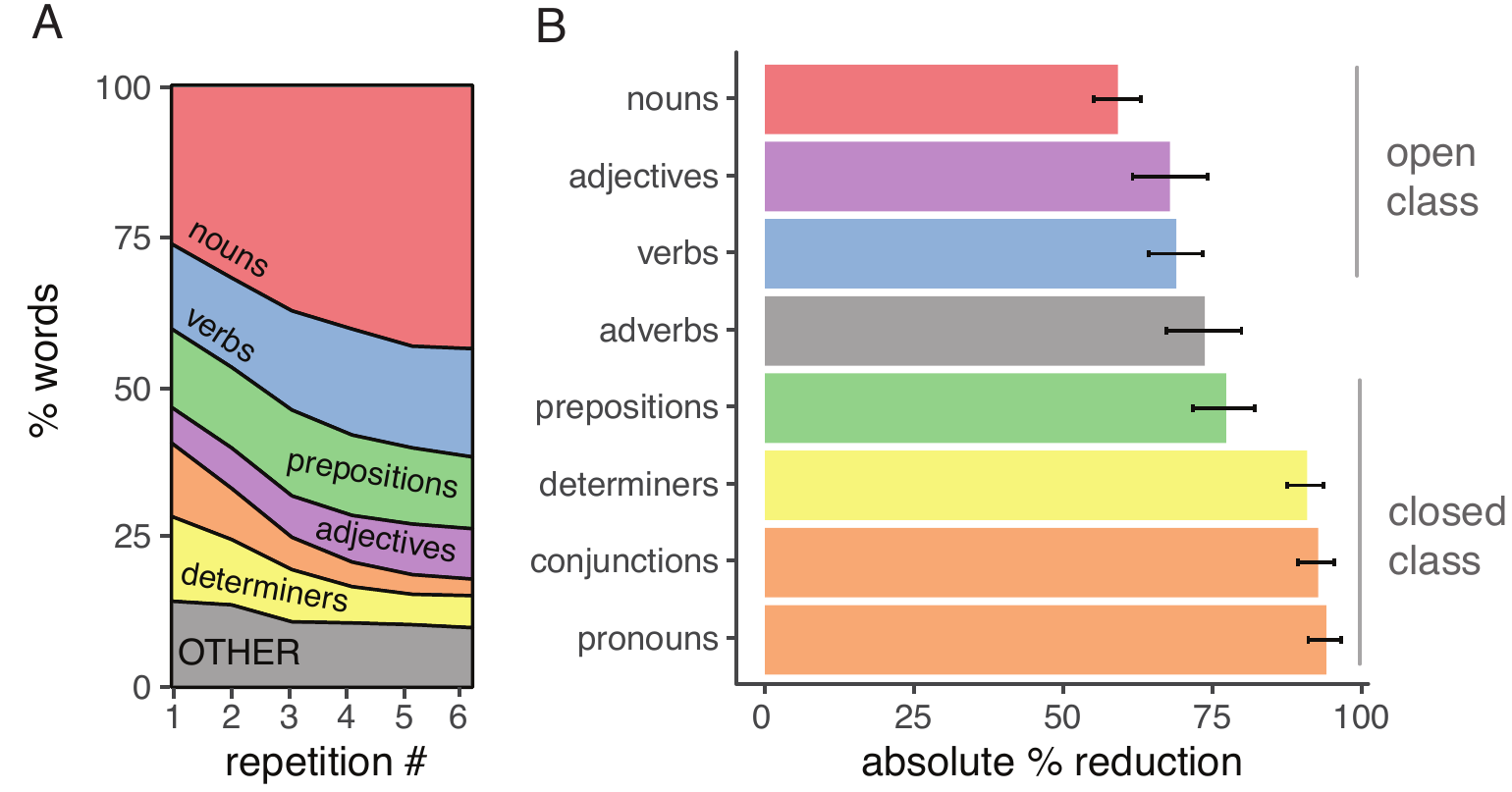}
\caption{(A) Proportion of words from different part of speech at each repetition block. For legibility, pronouns and conjunctions are combined in the orange strip while adverbs were grouped into ``OTHER''. (B) Closed-class parts of speech are more likely to be dropped than open-class parts of speech. Note that the classification of adverbs is controversial, as many common adverbs are considered closed-class items (e.g. ``only,'' ``now,'' ``there'') while others are open. Error bars are bootstrapped 95\% confidence intervals.}
\label{fig:pos}
\end{figure}

\begin{table*}[t]
\centering
\begin{tabular}{|r||l|l|l|}
  \hline
 & unigrams & bigrams & trigrams \\
  \hline
\#1 & a & look like & look like a \\
  \#2 & the & like a & look like -PRON- \\
  \#3 & -PRON- & to the & to the right \\
  \#4 & like & this one & like a person \\
  \#5 & look & the right & to the left \\
  \#6 & be & the left & one look like \\
  \#7 & on & like -PRON- & this one look \\
  \#8 & one & on the & like -PRON- be \\
  \#9 & with & with a & this one be \\
  \#10 & to & a person & -PRON- look like \\
  \#11 & and & -PRON- be & look like someone \\
  \#12 & right & on top & diamond on top \\
  \#13 & this & a diamond & in the air \\
  \#14 & of & in the & on top of \\
  \#15 & head & one look & a diamond on \\
   \hline
\end{tabular}
\caption{Top 15 unigrams, bigrams, and trigrams with the highest numeric reduction from first repetition to last repetition. Text lemmatized before n-grams computed, which also mapped all pronouns to the ``-PRON-'' token. }
\label{tab:words}
\end{table*}

\subsection{Breaking down the structure of reduction}\label{reduction}

\paragraph{Reduction in parts of speech}

Having established the matcher-dependent conditions under which directors are willing to shorten their utterances, we now examine the way they are shortened in more detail.
First, we explore which \emph{kinds} of words are most likely to be dropped.
We used the SpaCy part-of-speech tagger \citep{spacy2} to count the number of words belonging to different parts of speech in each message sent by the director\footnote{The SpaCy tagger is statistical, meaning it uses a neural network to return the (uniquely) most probable assignment of tags under its model. It obtains comparable accuracy ($\sim 97\%$) to other modern taggers \citep{manning2011part}. However, it is important to note that the language used in our task likely differs from the tagger's training sample, containing higher rates of sentence fragments, bare NPs, and `ungrammatical' language that human annotators might also find difficult to classify into standard parts of speech. The SpaCy dependency parser (see below) is similarly statistical and unlikely to be perfect. For example, neural parsers experience particular difficulty determining exactly where to attach conjunction relations \citep{ficler-goldberg-2017-improving}. As the state-of-the-art techniques for these tasks continue to improve, we hope future work using our corpus will be able to achieve more precise estimates.}.
In Fig. \ref{fig:pos}A, we show the shifting proportions of different parts of speech at each repetition.
We find that nouns account for proportionally more of the words being used over time, while determiners and prepositions account for fewer.

To test which kinds of words are more likely to be dropped, we measured the percent reduction in the number of words in each part of speech from the first repetition to the sixth repetition (i.e. $(n_1 - n_6)/n_1$).
We find that pronouns (`it', `he'), conjunctions (`and', `that'), and determiners (`the', `a', `an') are the most likely classes of words to be dropped (94\%, 93\% and 91\%, respectively) and nouns (`dancer', `rabbit') are the least likely to be dropped (59\%).
More generally, closed-class parts of speech, including function words, appear strictly more likely to be dropped than open-class parts of speech (Fig. \ref{fig:pos}B).
To statistically test this claim, we constructed a mixed-effects model predicting the reduction of each part of speech from the first round to the final round.
We included the binary fixed effect of open class vs. closed class (excluding adverbs), as well as random intercepts and slopes for each pair of participants, and random intercepts for each individual part of speech within each class.
We found a significant effect of class, $b=0.24, t = 4.5, p = 0.006$, supporting the observation that closed class parts of speech are more likely to be dropped. 

One possible interpretation of these findings is that reduction may be driven mostly by the loss of function words as directors shift to a less-grammatical shorthand over the course of the task.
However, when examining the n-grams most likely to be dropped (see Table \ref{tab:words}), we noticed that many of the most dropped closed-class words are used to  form prepositional phases (`of', `with') or combine different clauses (`and').
Others are modifiers (`the right \dots').
These examples suggest an alternative explanation: the higher reduction of closed-class function words may be a consequence of entire meaningful grammatical units (e.g.~clauses, prepositional phrases) being dropped at once.

\begin{figure}[t!]
\centering
\vspace{-1em}
\includegraphics[scale=.75]{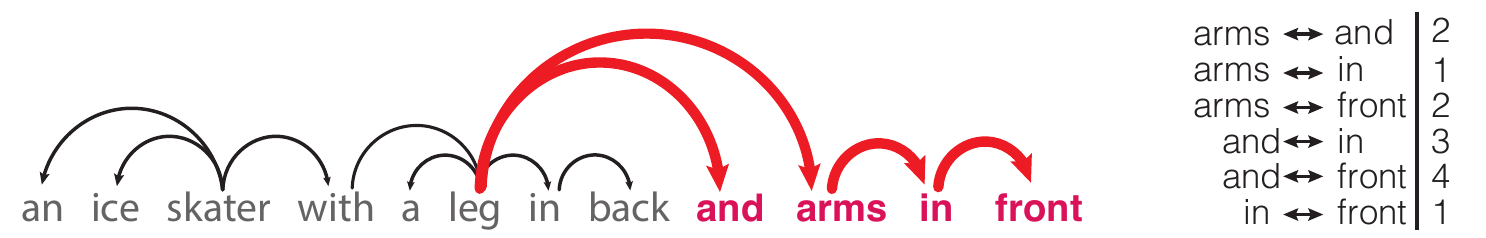}
\caption{Example dependency parse for a referring expression in our task. If the words ``and arms in front'' were dropped, we would find a mean path length of 2.17 among the dropped words.}
\label{fig:dependency}
\end{figure}

\paragraph{Reduction in syntactic units}
If initial descriptions tend to be syntactically complex because they combine multiple pieces of identifying information about the target, then the director may gradually omit meaningful `chunks' as they become informationally redundant.
We explicitly tested this hypothesis by examining whether pairs of words dropped from one reference to the next tend to come from the same syntactic units, relative to what would be expected by random deletions.
We quantified the extent to which dropped words `cluster' by examining closeness between the dropped words in a dependency parse tree (see Fig. \ref{fig:dependency}).
Dependency grammars represent syntax in terms of directed dependency relations between pairs of words \citep{hudson1984word,corbett1993heads}, and dependency parsing, the task of extracting a tree of dependency relations from raw input text, has become a canonical task in computational linguistics \citep{jurafsky2014speech,kubler2009dependency}\footnote{Dependency grammars are an alternative to phrase structure grammars, which represent \emph{constituency} relations between abstract units like noun phrases and verb phrases. We pre-registered a different version of this analysis using the pure prevalence of different higher-order clausal features extracted from a constituency parse at each repetition (e.g. adjectival clauses or subordinate clauses). While we found evidence for our pre-registered predictions, we later realized that this measure failed to capture the relevant notion of clustering: clausal features can disappear due to rephrasing without the words in the clause necessarily being dropped together, and conversely, entire meaningful units can be dropped at once (e.g. noun phrases) without any clausal features being present at all. Path lengths in the dependency tree are better suited for assessing these questions at the word-by-word level.}. 

We used the state-of-the-art statistical dependency parser from SpaCy, using the ClearNLP dependency labels \citep{choi_palmer2012}, to extract a parse tree for every utterance in our corpus \citep[see][for details on neural network architectures for parsing]{honnibal2015improved, kiperwasser2016simple}.
We then compared each referring expression to the one produced on the subsequent repetition block to determine which words were dropped and which reappeared.
For each each pair of words that were dropped, we used a standard shortest-path algorithm to find the minimal distance between them in the dependency parse tree we extracted (see Fig. \ref{fig:dependency}).
Finally, we computed the mean path lengths between all such pairs of dropped words on each given trial, and then took the mean across all trials (excluding blocks where no words were dropped).
This method weights each utterance evenly, preventing trials with more words from dominating the global average.

We compared this empirical `syntactic clustering' statistic to two baselines.
For the \emph{random} baseline, instead of examining dependency lengths between the words that were actually dropped, we randomly sampled the same number of words from the referring expression and computed the dependency length between them.
We repeated this procedure 100 times to obtain a null distribution of the mean dependency length that would be expected if words were being dropped \emph{randomly} from anywhere in the message.
For the \emph{function words} baseline, we were specifically interested in the null distribution that should be expected if function words were preferentially dropped independent of the syntactic sub-units they belong to.
We first sampled from the set of function words in the utterance, and if this set was smaller than the total number of words dropped, we filled the remainder with random non-function words.

We found a mean empirical dependency length of 2.77, which lay outside the both the random null distribution (range: $[2.90, 2.99]$) and the function word null distribution (range: $[3.03, 3.08]$), indicating a reliable effect of syntactic clustering among the words that were dropped on each round.
That is, these words tended to be closer to one another in the dependency parse than expected by total chance or by preferentially dropping function words independently of their corresponding syntactic units.
Furthermore, while overall dependency lengths get smaller as utterances become shorter, this result holds within every repetition block (see Supplemental Fig. \ref{fig:dependency_length_by_rep}), and other statistics gave similar results, including the minimum dependency length and the raw distance in the sequence of words.
This result accords with earlier observations by \cite{Carroll80_NamingHedges}, who reanalyzed transcripts from \cite{KraussWeinheimer64_ReferencePhrases}. 
In those data, the short names that participants converged upon were prominent in some syntactic construction at the beginning of the session, often as a head noun that was initially modified or qualified by other information.

\section{Results: characterizing the dynamics of semantic content}
\label{sec:content}

So far we have examined the increasing efficiency of referring expressions in terms of their (syntactic) structure.
We next explore how the \emph{semantic content} of referring expressions changes over repeated reference.
Which words from a speaker's initial description are most likely to become conventionalized in their final labels?
Why do all dyads not end up with the same conventions? And,
once efficient conventions are formed, are they stable?
In exploring these questions, we find support for a view of adaptation as a path-dependent process of gradually paring down redundant information and coalescing around the most diagnostic features for the given context.

\subsection{Initially distinctive words are more likely to conventionalize}
\label{sec:distinctive}

Two general computational principles guide our exploration of which content is dropped and which is preserved.
First, Gricean principles suggest that a good referring expression is one that applies more strongly to the target than to the distractors; in contrast, those expressions that apply to multiple objects will be less informative.
Second, principles of cross-situational learning suggest that these informativity considerations will be strengthened over time.
The exclusive usage of a word with one tangram and no others should reinforce the specificity of that meaning in the local discourse context, even if the matcher may be \emph{a priori} willing to extend it to other targets.
Conversely, if a particular word has been successfully used with several different referents, its specificity may be weakened in the local context.
Putting these principles together, we hypothesized that the labels that conventionalize should not be a random draw from the initial description.
Instead, more \emph{initially distinctive} words should be more likely to conventionalize.

For each pair of participants, we quantified the distinctiveness of a word $w$ as $n_w$: the number of tangrams that it was used to describe on the first repetition.
A word that is only used in the description of a single tangram (e.g. a descriptive noun like ``rabbit'') would be very distinctive, while a word used with all 12 tangrams (e.g. an article like ``the'') would be not distinctive at all.
While this formulation is easy to state in words, it is equivalent (up to a simple deterministic transformation) to two popular and theoretically motivated measures of distinctiveness used in natural language processing \citep{salton1988term}: \emph{tf-idf} and \emph{PPMI}.\footnote{
The first is \emph{term frequency-inverse document frequency} \citep[tf-idf,][]{sparck1972statistical}, which multiplies the term frequency $tf(w,d)$ of a word $w$ in a document $d$ by a ``global'' term $\log(N/n_w)$ where $N$ is the total number of documents and $n_w$ is the number of documents containing $w$.
In our case, the ``documents'' are just the referring expressions used for a distinct tangram on the first repetition, so $N=12$ and we can take $tf(w,d)$ to be a boolean for simplicity: 1 if the word occurs, 0 if it does not.
We can thus retrieve our simpler measure by exponentiating, dividing by $12$, and taking the inverse.
The second is \emph{positive point-wise mutual information} (PPMI).
Point-wise mutual information compares the joint probability of a word occurring with a particular tangram to the probability of the two occurring independently:
$$PMI_{\textrm{word},\textrm{tangram}} = \log\frac{P(\textrm{word}, \textrm{tangram})}{P(\textrm{word})P(\textrm{tangram})}$$
\emph{Positive} point-wise mutual information is given by $\min(0, \textrm{PMI})$, restricting the lower bound to 0.
It can be shown for our case that \emph{tf-idf} is the maximum likelihood estimator for PPMI: the numerator reduces to a boolean when we only have one observation per tangram \citep{robertson2004understanding}. Our pre-registered analysis plan used PPMI, and our result holds using either measure; however, we decided to report the pure tangram count measure as more interpretable.}
\begin{figure}[t!]
\centering
\includegraphics[scale=.85]{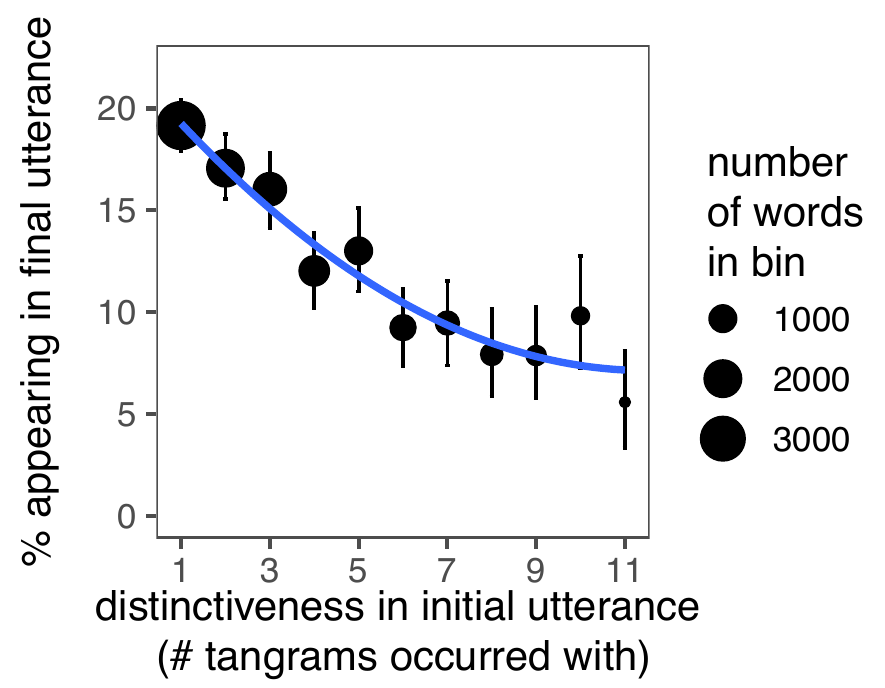}
\caption{More distinctive words are more likely to conventionalize. Points represent estimates of the mean probability of conventionalizing across all words with a given distinctiveness value. Size of points represent the number of words at that value. Curve shows regression fit; error bars are bootstrapped 95\% CIs.}
\label{fig:distinct}
\end{figure}
Given this simple but principled measure of word distinctiveness at the speaker-by-speaker level, we were interested in the extent to which it accounts for conventionalization: the probability that a word in the director's initial description is preserved until the end of the game.
More than half of the words used to refer to a tangram on the final repetition (57\%) appeared in the initial utterance.\footnote{The 43\% of final repetition words that did not exactly match were sometimes synonyms or otherwise semantically related to words used on the first repetition, e.g. ``foot'' on the first repetition vs. ``leg'' on the last. In other cases, the labels used at the end were introduced after the first repetition, e.g. one pair only started using the conventionalized label ``portrait'' on repetition 3.}
We thus restricted our attention to this subset of words, coding them with a 1 if they later appeared at the final repetition and 0 if they did not.
We then ran a mixed-effects logistic regression including a fixed effect of initial distinctiveness and maximal random effect structure with intercepts and slopes for each tangram and pair of participants.
We found a significant positive effect of distinctiveness: words that were used with a larger number of tangrams on the first repetition were less likely to conventionalize, $b = -0.23, z = -6.1$ (see Fig. \ref{fig:distinct}).
Similar results are found explicitly using the \emph{tf-idf} measure.

To further evaluate how distinctiveness is related to eventual conventionalization, we conducted a non-parametric permutation test.
For each speaker and tangram, we extracted the word with the \emph{highest distinctiveness value} and computed the mean probability of this word also being used on the final repetition. 
In the case of a tie, we sampled from the set of words sharing the highest distinctiveness value.
After repeating this procedure 1000 times, we found a distribution ranging from 24\% to 31\%.
As a baseline null model, we randomly sampled from the list of all words contained in the initial utterance instead of the most distinctive one.
Repeating this procedure yielded a null distribution ranging from 2.5\% to 6.6\%, which was significantly lower than the one derived from the most distinctive words.
These results are also consistent with our earlier finding that open-class parts of speech are more likely to be preserved on the final repetition than closed-class parts of speech (see Section \ref{reduction}), since open-class parts of speech are statistically more likely to supply distinctive words. 

\subsection{Semantic meaning diverges across pairs and stabilizes within pairs}

Conventions are characterized by their arbitrariness and stability \citep{Lewis69_Convention}.
Our remaining predictions concern the dynamics of these properties.
First, due to sources of variability in the population of speakers, we predict that the referring expressions used by different pairs will increasingly diverge to different, idiosyncratic labels.
In other words, different pairs will find different but equally successful equilibria in the space of possible linguistic conventions.
Second, as directors learn and gradually strengthen their expectations about how their partner will interpret their referring expressions, the labels used within each pair for each tangram will stabilize.
In other words, once there is evidence that a particular label is successfully understood, there should be little reason to deviate from it.

\begin{figure}[t!]
\centering
\includegraphics[scale=.8]{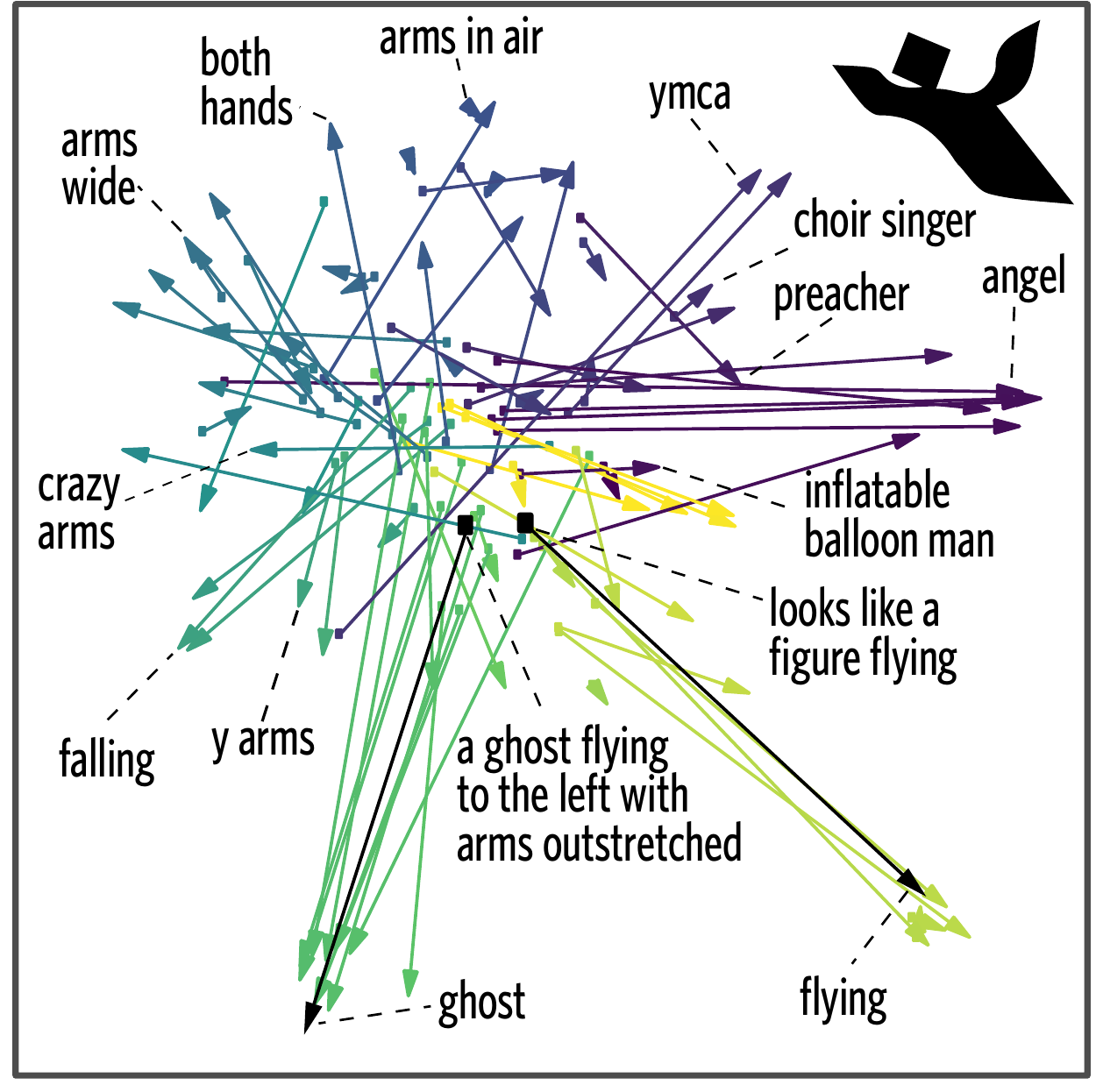}
\caption{2D projection of semantic embeddings for example tangram using t-SNE. Each arrow represents the trajectory between the first repetition to last repetition for a distinct pair of participants. Color represents the rotational angle of the final location to more easily see where each pair began. Annotations are provided for select utterances, representing different equilibria found by different participants. Arrows in black highlight a pair of trajectories where the initial utterances were similar but the final equilibria were differentiated. Because t-SNE is a stochastic algorithm, even identical words (e.g. the many instances of ``ghost'') will map to slightly different locations.}
\label{fig:tsne}
\end{figure}

To operationalize these constructs, we used a measure of similarity based on distances computed between continuous vector space embeddings of referring expressions\footnote{We pre-registered discrete analogs of these analyses, which we have included in Appendix A, but we later decided that the continuous vector space measures present a more interpretable and direct test of our hypothesis.}.
Although the idea of using such representations of words to measure similarity is an old one \citep{osgood1952nature,landauer_solution_1997,bengio_neural_2003}, recent progress in machine learning has yielded substantial improvements in the quality of these representations.
To quantify the dynamics of semantic content in referring expressions across and within games, we therefore first extracted the 300-dimensional GloVe vector \citep{pennington2014glove} for each word in the messages produced by the director on each trial.
We limited our analysis to messages produced by the director, rather than collapsing together director and matcher messages, in order to preserve consistency in the source of semantic content.
We then averaged these word vectors to obtain a single sentence vector for each trial.
Variations on such simple averaging methods are surprisingly strong baselines for sentence representations \citep{arora2017asimple}, providing better downstream task performance than whole-sentence encoders \citep{KirosEtAl15_SkipThought}.

To avoid artifacts from function words, we only included open-class content words (nouns, adjectives, verbs, and adverbs) in this average
Because different forms of a word may have slightly different representation, we also applied a lemmatizer to further standardize the input. Lemmatization maps multiple morphological variants (e.g. `played,' `playing,' `plays') to the same stem (`play'). 
We did not want an observed difference between two pairs to be driven simply by different forms of the same word.
We then defined a similarity metric between any pair of sentence vectors $\langle u_i, u_j \rangle$.
Our results are robust to several choices of metric, but for simplicity we will use cosine similarity throughout the presentation below: $$\langle u_i, u_j \rangle = \cos \theta_{ij} = \frac{u_i \cdot u_j}{\| u_i\| \| u_j \|}$$

We begin by visualizing the trajectories taken by each pair of participants when referring to a particular example tangram (see Supplemental Figure \ref{fig:all-tsne} for similar plots for the other items).
To create this visualization, we took the first 50 components recovered by running Principal Components Analysis (PCA) on the 300-dimensional embeddings for all utterances used to refer to this tangram, including all speakers and all repetition blocks.
We then used t-SNE \citep{maaten2008visualizing} to stochastically embed the lower-dimensional PCA representation of these utterances in a common 2D vector space\footnote{t-SNE is a stochastic, non-linear dimensionality reduction technique which focuses on keeping neighboring points in the high-dimensional space close together in the lower-dimensional space. An initial linear reduction to an intermediate dimensionality is commonly used to speed up computation and reduce noise in the high-dimensional space, compared to applying t-SNE directly to the 300-dimensional vectors. Conversely, the advantage of using t-SNE over projecting directly to 2-dimensions with a linear technique like PCA is its ability to preserve non-linear structure in the high-dimensional space.}.
Finally, we connected the first and last utterance a particular pair used to refer to this tangram with an arrow (Fig. \ref{fig:tsne}), and annotated utterances in several regions of the space.

Most strikingly, we observed that the initial utterances of each game tend to cluster tightly near the center of the space and the final utterances are \emph{dispersed} more widely around the edges.
This pattern is consistent with the hypothesis that early descriptions may overlap before each speaker hones in on more distinctive different equilibria later in the game.
Indeed, pairs often initially mentioned multiple properties (e.g. ``person raising their arms up like a choir singer'') before breaking the symmetry and collapsing to one of these properties (``choir singer'').
Our example also shows the variety of different solutions discovered by different speakers.
A handful of semantically distinct labels served as equilibria for a number of pairs (``ghost,'' ``flying,'' ``angel'') while many more idiosyncratic labels spread out more widely in space.
In the remainder of this section, we test these observations.

\begin{figure}[t!]
\centering
\includegraphics[scale=.75]{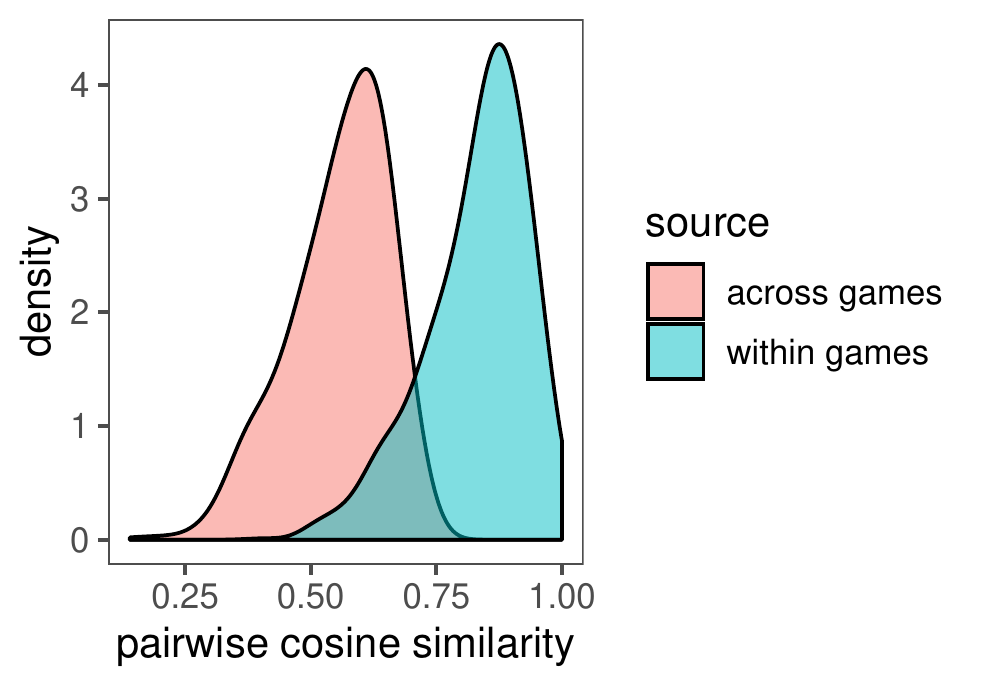}
\caption{Distribution of similarities between different utterances within and across different games.}
\label{fig:withinvsacross}
\end{figure}

\paragraph{Utterances are more similar overall within games than between games}
Before examining the dynamics of how these vectors change over time, we test the basic prediction that referring expressions used by a single speaker \emph{within} a game are more similar overall than those used by different speakers \emph{across} games.
For each tangram, we computed the pairwise similarities between all utterances used by a speaker to refer to that tangram at different times \emph{within} a game and also between all utterances used by different speakers \emph{across} games.
The distributions of these values are shown in Fig. \ref{fig:withinvsacross}.
We estimated the distance between these distributions using the standard normalized sensitivity $d' = \frac{\mu_A - \mu_W}{\sqrt{1/2(\sigma^2_A+\sigma^2_W)}} = 2.65$.
To compare this estimated difference against the null hypotheses that within- and across-game similarities are drawn from the same distribution, we conducted a permutation test by scrambling `within' and `across' labels for each similarity and re-computing $d'$ 1000 times.
We found that our observed value was extremely unlikely under this null distribution, $95\%~CI: [-0.09, 0.09], p < 0.001$.
In other words, utterances from a single pair tend to cluster together in semantic space while different pairs are spread out in different parts of the space.
This observation leaves open the question of whether pairs start out semantically similar and become different through the conventionalization process (as predicted by the theory of conventions), or simply come into the experiment with idiosyncratic differences.
To explore this question, we conducted analyses on how the semantic vectors changed over time.

\begin{figure}[t]
\centering
\includegraphics[scale=.5]{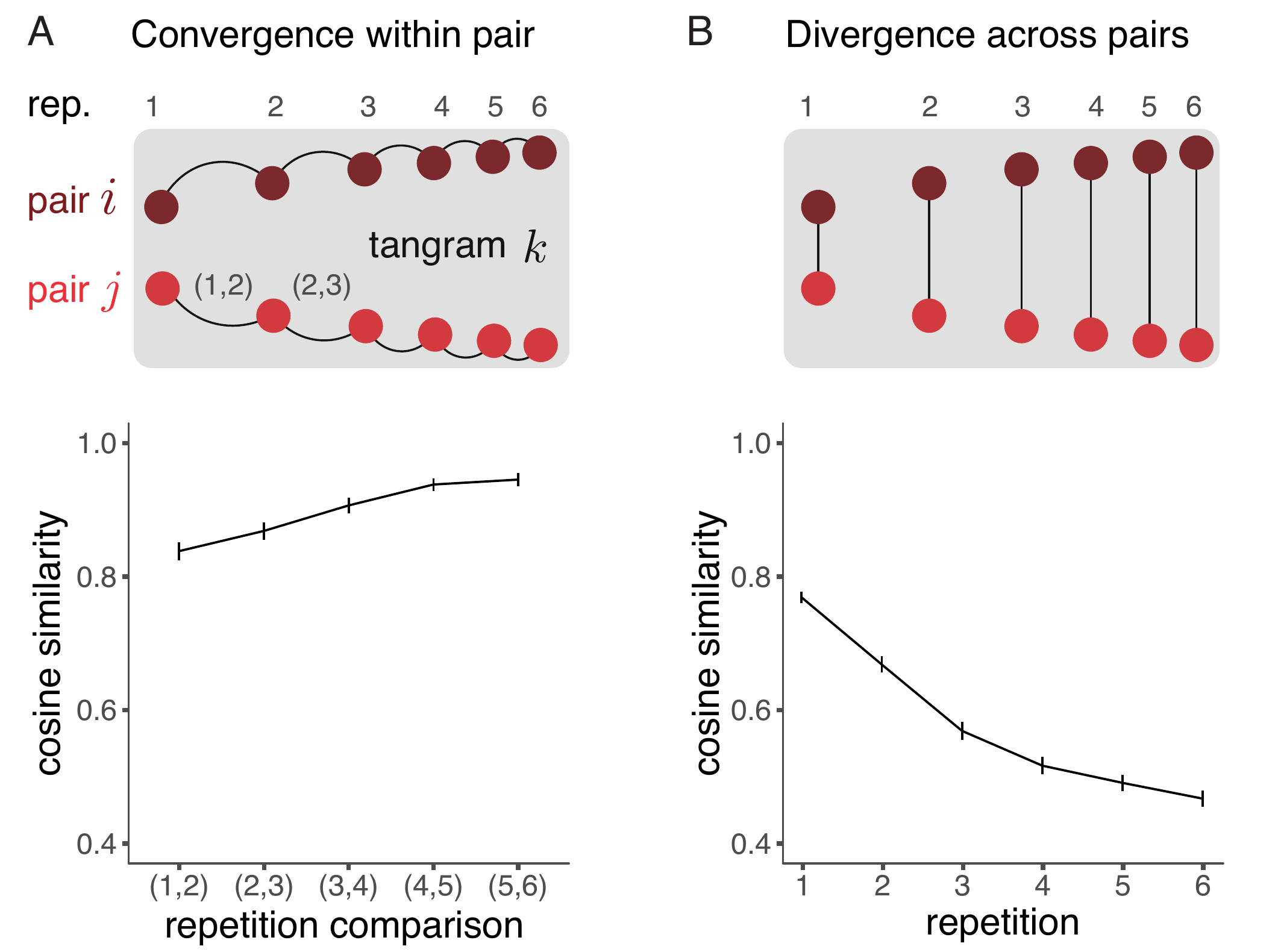}
\vspace{1em}
\caption{(A) Utterances within a pair become more similar to successive utterances on later repetitions, converging on a stable convention, but (B) utterances across pairs become steadily more dissimilar, diverging to different solutions. These patterns are depicted schematically by dots within a pair changing less over time while dots in different pairs move further apart. Error bars are bootstrapped 95\% CIs over participant-level means.}
\label{fig:similarity}
\end{figure}

\paragraph{Utterances become increasingly consistent within interaction}
As directors modified their utterances across successive repetitions, we hypothesized that they would converge on increasingly consistent, stable ways of referring to each tangram.
To test this prediction, we computed the cosine similarity between successive utterances produced by each speaker (see Fig.~\ref{fig:similarity}A). 
A mixed-effects model with (orthogonalized) linear and quadratic fixed effects of repetition number and maximal random effects for both tangram and pair of participants showed that similarity between successive utterances increased substantially throughout an interaction ($b = 2.8,~t = 11.0, p < 0.001$).
The quadratic term was also weakly significant ($b= -0.44,~t=-2.2, p = 0.035$).


\paragraph{Utterances become increasingly different across interactions}

Finally, we predicted that although the referring expressions used by different pairs may begin with substantial overlap, they would become increasingly dissimilar from each other across time, gradually diverging into different equilibria.
We tested this prediction by computing the mean similarity between referring expressions used by different speakers.
The large sample of similarities ($N = 245,016 = 12~\texttt{tangrams} \times 6~\texttt{repetitions} \times \frac{83 \cdot 82}{2}~\texttt{distinct pairs}$) presented both advantages and disadvantages for this analysis.
On one hand, we could obtain highly reliable estimates of mean similarity.
On the other hand, larger random-effects structures led to convergence problems.
We therefore ran a mixed-effects regression model including linear and quadratic fixed effects of repetition number including random effects only at the tangram-level.
We found a strong negative linear fixed effect of repetition on between-game semantic similarity ($b = -48.6, t= -16.7, p <0.001$) as well as a significant quadratic effect ($b= 15.2, t = 10.9$), indicating that this divergence slows over time (likely due to stabilization within interactions; see Fig.~\ref{fig:similarity}B).


\section{General Discussion}
\label{sec:discussion}

Our language changes as we get to know a social partner through repeated interactions.
We gradually learn what is meaningful to them and establish common ground.
In this paper, we characterized the quantitative dynamics of this process by examining behavior in a new corpus of natural-language repeated reference games.
This corpus is sufficiently large to provide new traction toward resolving theoretical questions about the nature of adaptation in communication. Our study illustrates the general point that larger datasets enable more precise measurements, which in turn drive theory development \citep{frank2018great}.


In our corpus, we replicate the classic finding that directors reduce the length of their descriptions over the course of the task. But we also show that they do so in a way that is sensitive to evidence of matcher understanding and structured to omit redundant syntactic chunks of information, leaving eventually only the most distinguishing words.
The resulting labels display quantitative signatures of increasing \emph{arbitrariness} in the sense that different pairs increasingly diverge to distinct solutions, and \emph{stability} in the sense that speakers do not deviate from a solution once it is discovered.
Taken together, these findings clarify the desiderata for theories of \emph{ad hoc convention formation}.
For a model of communication to explain how general-purpose meanings are systematically tailored to the needs of the current interaction in the way we observed, it must provide a mechanism to select and combine syntactic phrases that are initially distinctive and to prune them over time, modifying them if they are unsuccessful.

Our findings also raise new and subtle questions about the cognitive mechanisms giving rise to these properties.
One such question concerns the mechanisms supporting arbitrariness over such short interactions: what breaks the `symmetry' among different possible descriptions and leads different pairs to diverge from one another?
One possibility is that each individual speaker may initially have strong but idiosyncratic initial preferences for short labels, and arbitrariness emerges from variability in these preferences throughout the population.
Under this possibility, speakers begin with long, elaborated descriptions due to uncertainty about whether their preferred label will be understood, but in the absence of misunderstandings will proceed with their pre-meditated label.
A second possibility is that speakers \emph{themselves} may be unclear about a mutually understandable way to refer to these unfamiliar objects.
If uncertain speakers initially sample an utterance from a broad distribution of acceptable labels, and update their distribution on subsequent repetitions conditioned on feedback, different pairs may end up in different equilibria due to randomness in sampling from a more or less shared initial distribution. 
This latter mechanism has been proposed in recent probabilistic models of convention formation \citep{smith_learning_2013,hawkins_convention-formation_2017,Brochagen17}, which present simulations reproducing several of the properties we observed in our data.
These two possibilities -- strong but idiosyncratic initial preferences or initial uncertainty and breadth -- are not mutually exclusive.
Our results rule out the possibility of universally shared strong preferences, but it is possible that some speakers have different strong preferences about labels while others are initially more uncertain.
One way for future work to disentangle these possibilities is to elicit better measurements of speakers' initial beliefs about appropriate labels.
For instance, an approach proposed by \cite{FussellKrauss89_IntendedAudienceCommonGround} asked directors to either produce descriptions for others or for themselves in the future, and Bayesian truth serum approaches \citep{prelec2004bayesian} compare an individual's own subjective preferences with their expectations about whether these would be shared by others.

Although our analyses go beyond previous work by using recent vector-space semantic models, they still face several limitations.
We address two such limitations with supplemental analyses included in the Appendix.
First, measures of similarity relying on vector space representations like GloVe are fundamentally limited by the quality of the semantic space that has been learned.
To address this concern, we provide converging evidence for the properties of arbitrariness and stability using the \emph{discrete distributions} of word tokens appearing in each utterance instead of continuous utterance embeddings (see Appendix A).
Second, a related concern is that the gradual divergence we observed between different interactions could be an artifact of the way we constructed utterance embeddings by averaging word embeddings.
If averaging together more words creates a distinctive type of `washed out' utterance embedding, and early descriptions contain more words, then high initial similarity across interactions may reflect utterance length rather than actual semantic overlap.
We address this concern by providing an additional permutation test baseline that scrambles words across different tangrams prior to averaging word embeddings (see Appendix B).
This baseline also presents an opportunity to compare the divergence we observed across different interactions against the divergence within a \emph{single speaker's} descriptions of their twelve different tangrams.
Just as different speakers initially include many of the same attributes in their descriptions for a tangram but eventually (unknowingly) diverge to distinct labels, a single speaker also begins by re-using certain attributes for several tangrams but (knowingly) prunes them down to be as distinctive as possible due to informativity pressures, consistent with our findings in Section \ref{sec:distinctive}.


Our use of classic tangram stimuli also raises an important question about how our findings would be expected to apply to other spaces of novel objects.
In particular, it is likely that participants converge to distinctive `names' because the target of reference were distinctive objects.
If the targets of communication instead varied along clear latent dimensions \citep[e.g.][]{nolle2018emergence}, contained multiple objects in relation to one another, or depicted events or activities unfolding over time, participants might instead have converged on more compositional systems making use of adjectives, verbs, and prepositions.
Similarly, the generality of our results is limited by the population we sampled.
Our use of online data collection allowed us to create a relatively large number of arbitrary dyads within a convenience population, but also limits opportunities for studying these dyads over longer time periods.
It will be important to determine how the \emph{ad hoc} meanings formed in one novel context generalize to other contexts with the same partner.
Further, though our dyads are likely diverse in many ways relative to the US national population \citep{levay2016demographic} they not representative of either the US population or any broader population.
Thus, further cross-cultural work examining the validity of our conclusions across populations, and in different languages, would be a valuable contribution for future work.

Finally, there are important differences between face-to-face verbal communication and online text-mediated communication.
Text message interfaces do not transmit the message character-by-character as it is being written; messages are seen in their entirety upon being sent. 
Meanwhile, speech is processed incrementally, allowing for a real-time ``backchannel'' where the director may be interrupted as soon as the matcher is confident they understand.
The timing of verbal interruptions may provide stronger evidence of understanding than a reply to a fully-composed message.
Still, we believe it is a strength of our study to collaborative reference through a text-based chat modality which is an increasingly common site of communication in the real-world. 
All of the key results from classic work in face-to-face verbal conversations were strongly replicated in our text-based data, suggesting that these processes are modality-general features of communication. 
Moreover, while a simple heuristic to keep talking until being interrupted could in principle suffice to explain speaker reduction face-to-face conversation, our evidence shows that speakers \emph{plan} shorter utterances even when they cannot be interrupted; social feedback is being used as latent evidence of understanding.
While our analyses of referential content were limited to those produced by the director to preserve the consistency of their source, it is important for future work to more systematically examine the matcher's contributions to establishing reference. 
For instance, in some cases the matcher may supply the label that goes on to be conventionalized, and may systematically produce different discourse acts in different contexts.

The rapid timescale of adaptation we have investigated in dyadic reference games is not only of interest in its own right for theories of meaning and social coordination; it is a key building block toward grounding the adaptiveness and efficiency of larger-scale human language in the cognitive mechanisms of individual minds \citep{KirbyTamarizCornishSmith15_CompressionCommunication,gibson2019efficiency}.
If community-wide conventions emerge from agents generalizing across different dyadic interactions, then local learning mechanisms leading to efficiency and informativity \emph{within} a dyad may explain how a community's conventions remain well-calibrated to the demands of the current environment.
The sensitivity of such calibration has been previous tested using small artificial languages in the lab \cite[e.g.][]{WintersKirbySmith14_LanguagesAdapt}, but our observation of similar dynamics in ordinary natural language use emphasizes that local learning may be an ongoing and pervasive influence.
In sum, the resolution provided by the larger corpus we have collected, in combination with recent advances in natural language processing techniques, provides a new window into the quantitative dynamics of adaptation in dyadic communication and beyond.
We hope that both the new corpus and new analytic techniques contribute to the testing and elaboration of theories of human language.

%
\bibliography{library}

\section*{Appendix A: Discrete word distributions}

Here we examine an alternative approach to evaluating claims of arbitrariness
 and stability using discrete word distributions instead of the continuous vector space measure used in the main text.
We begin by examining the discrete \emph{distribution of words} that each pair uses to refer to each tangram, excluding stop words.
This distribution is a \emph{unigram} distribution over the vector of words that appear throughout the utterances produced by a given speaker to refer to a particular object (the modal support size of this distribution is 7 words.)
If a pair of participants converges on stable labels for a tangram, this stability should manifest in a highly structured distribution over words throughout the game for that pair.
If different speakers discover diverging conventions, this idiosyncracy should manifest in differing word distributions.
We formalize these intuitions by examining entropy, an information-theoretic measure: $$H(W) = \sum_w P(w) \log P(w)$$
The entropy of the word distribution for a pair is maximized when all words are used equally often and declines as the distribution becomes more structured, i.e.~when the probability mass is more concentrated on a subset of words.\footnote{It also increases as a function of the support size; because in principle we consider this an important signature of a game, we focus on this unnormalized measure; however, the results hold if we control for the support size (i.e. divide the entropy by $\log(N)$ so that a uniform distribution will always have the maximum value of one.)}

To compare word distributions across games, we use a permutation test methodology.
By scrambling referring expressions for each tangram across games and recomputing the entropy of the scrambled word distribution, we effectively disrupt any structure within each pair.
There are two important inferences we can draw from this test.
First, in a null scenario where different pairs did \emph{not} diverge as predicted and instead every pair coordinated on roughly the same (optimal) convention for each tangram, this permutation operation would have no effect since it would be mixing together copies of the same distribution.
Second, in another null scenario where pairs did not converge and instead varied wildly in the words they used from repetition to repetition, then permuting across games would also have no effect since it would simply mix together word distributions that already have high entropy.
Hence, scrambling should \emph{increase} the average game's entropy only in the case where both predictions hold: each game's idiosyncratic but concentrated distribution of words would be mixed together to form more heterogeneous and therefore high-entropy distributions.

Following this logic, we computed the average within-game entropy for 1000 different permutations of director utterances.
We permuted utterances within repetition blocks rather than across the entire data set to control for the fact that earlier trials may generically differ from later ones (e.g. in utterance length).
Because we are permuting and measuring entropy at the tangram-level, this yields 12 permuted distributions (see Supplementary Fig. \ref{fig:permuted}).
We found that the mean empirical entropy lay well outside the null distribution for all twelve tangrams, $p < .001$, consistent with our predictions of internal stability within pairs and multiple equilibria across pairs.

Finally, it is worth noting some advantages and disadvantages of this discrete measure compared to the continuous vector space measure used in the main text.
A key advantage is that the entropy is not dependent on any particular choice of pre-trained vector embedding.
Due to biases in the vocabulary of their training corpora, vector representations also may not capture some of the more idiosyncratic conventions that participants converge on (e.g.  ``zig zag'' or ``Frank'' -- short for ``Frankenstein'').
Thus, to the extent we find converging results, the discrete measure may address concerns about the quality of the continuous representation.
A key disadvantage, on the other hand, is that our permutation test methodology is more indirect and does not have a natural scale.
We can strongly reject the complete absence of arbitrariness and stability --- a lower bound --- but there is no clear derivation for a corresponding upper bound showing exactly how strong these effects are.
Directly measuring divergence between word distributions is technically possible using divergence measures, but would not be informative at the fine granularity required for these analyses (i.e. at the level of single utterances).
Most utterances use entirely disjoint sets of words, and on later repetitions, the distribution may only contain one or two contentful words.
A final disadvantage is that discrete analyses treat even close synonyms as entirely distinct tokens in the word distribution because they are based entirely on the frequency of tokens rather than semantic content.
In summary, these two approaches provide complementary and converging evidence.

\section*{Appendix B: Additional baselines for evaluating divergence}

Could the divergence effect reported in section 3.2 be explained away as an artifact of our procedure for computing utterance embeddings?
If averaging together greater numbers of word vectors generically causes the resulting utterance vectors to be washed out and more similar one another, then the decrease in semantic similarity could be explained by a decrease in utterance \emph{length} (see Section 4.2) rather than divergence in content.
We tested this null hypothesis using a further  permutation test.
We reasoned that if the effect is in fact driven by length, then the similarity measured across interactions should be invariant to re-sampling utterance \emph{content}---the individual words that will be averaged together---within interactions.
We thus scrambled the words used by a participant across all twelve tangrams at each repetition, destroying any tangram-specific semantic content, but preserving utterance length.
By repeating this procedure 100 times, we found that the true mean similarity across pairs was higher than predicted under the null distribution at all six repetitions, $p < 0.01$, suggesting that the divergence effect is not solely driven by utterance length.

At the same time, we observed that this permutation test disrupted the mean similarity less than expected.
On the first repetition, for instance, the range of the null samples was $[0.754, 0.764]$, only slightly lower (in absolute terms) than the empirical value of $0.774$.
Why would this be the case, and how should we interpret the absolute degree of divergence?
One possibility is that there is already substantial semantic overlap on the first round in how a single speaker refers to different tangrams, so that scrambling does not dramatically disrupt the semantic content.
This possibility suggests examining the divergence between utterances used to refer to \emph{different tangrams within an interaction} as a useful baseline.
Based on our results in Section 3.1, we predicted that pragmatic pressures would lead labels for different tangrams \emph{within} an interaction to diverge more strongly than those for the same tangram \emph{across} interactions, despite starting with roughly similar overlap.
Indeed, we found that the average semantic similarity within an interaction was indistinguishable from the similarity across interactions on the first repetition (\emph{paired difference}: $0.003$) but the gap appeared to widen over subsequent rounds, indicating that the pressure to distinguish tangrams leads to greater divergence for a single speaker than the neutral divergence across different speakers would predict.

To test the statistical significance of this observation, we conducted a model comparison between mixed-effects models.
The dependent variable in both models is the difference score between mean within-speaker and across-speaker similarities (aggregated at the level of the speaker).
In the null model, we include only an intercept, which allows for a non-zero difference but does not allow this difference to increase or decrease with time.
In the full model, we additionally include a linear term for repetition number.
Because we have a mean difference score for each speaker, we also include random intercepts at the speaker level for both models.
A likelihood ratio test between these models shows that the full model fits the data significantly better, controlling for the additional degree of freedom, $\chi^2(1) = 10.8, p < 0.001$.\footnote{We have focused on this comparison to hold random effects constant, but including an additional term for a quadratic effect of repetition and an additional random effect of repetition are also supported by likelihood ratio tests. In this full model, we find a marginally significant linear effect of repetition, $b = 0.15, t = 1.9, p = 0.059$.}

To summarize, we suggested the use of a baseline to better interpret our core result showing divergence in labels across different speakers as pairs discover different conventions.
This baseline---the divergence in a speaker's \emph{own} utterances for different tangrams---begins at a similar level, indicating that the initial utterances used by different speakers overlap approximately as much as the different initial utterances used by a single speaker.
While both subsequent trajectories indicate divergence, the different labels used by a single speaker rapidly spread out in vector space and become more distinct from one another than the labels used by different speakers.

\newpage

\section*{Appendix C: Supplemental figures}

\begin{figure}[h!]
\centering
\includegraphics[scale=.9]{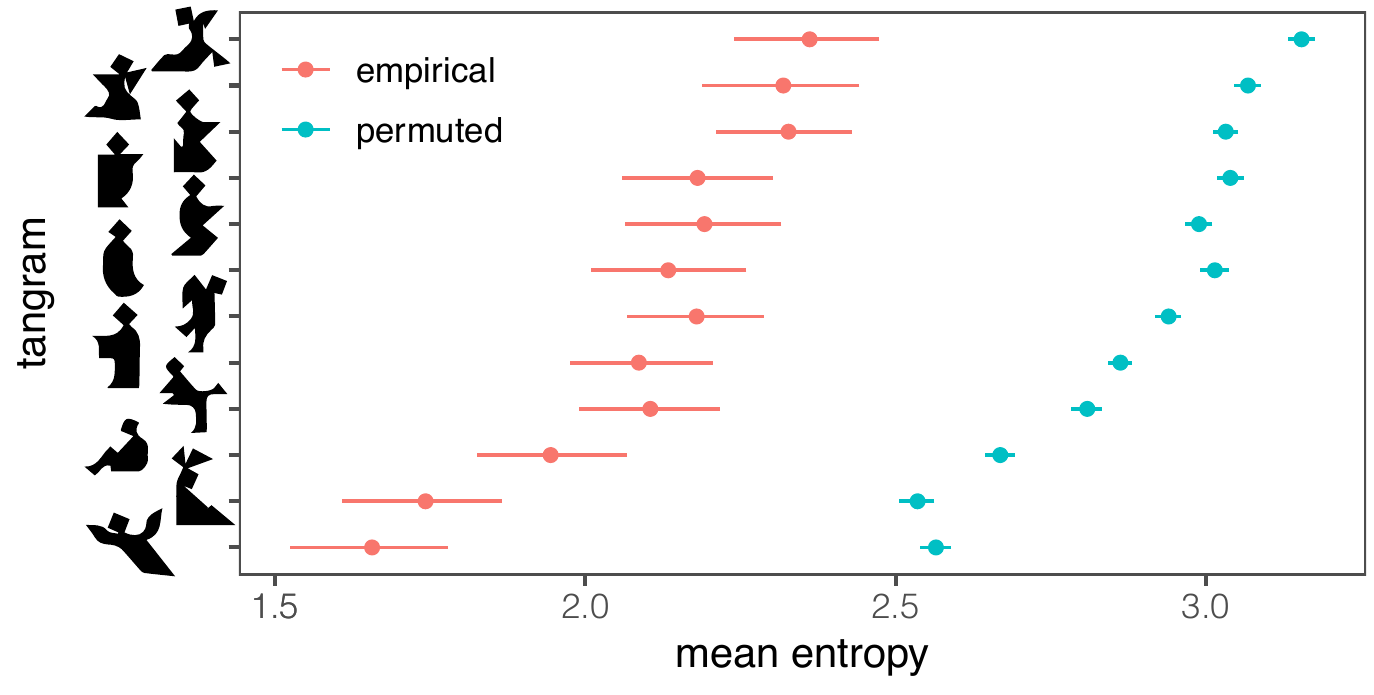}
\caption{Permuting utterances across pairs increases entropy of word distribution, consistent with internal stability and multiple equilibria. Mean empirical entropy (red) and mean permuted entropy (blue) are shown for each tangram. Error bars are 95\% CIs for bootstrapped empirical entropy and the permuted distribution, respectively.}
\label{fig:permuted}
\end{figure}

\begin{figure}[h!]
\centering
\includegraphics[scale=.8]{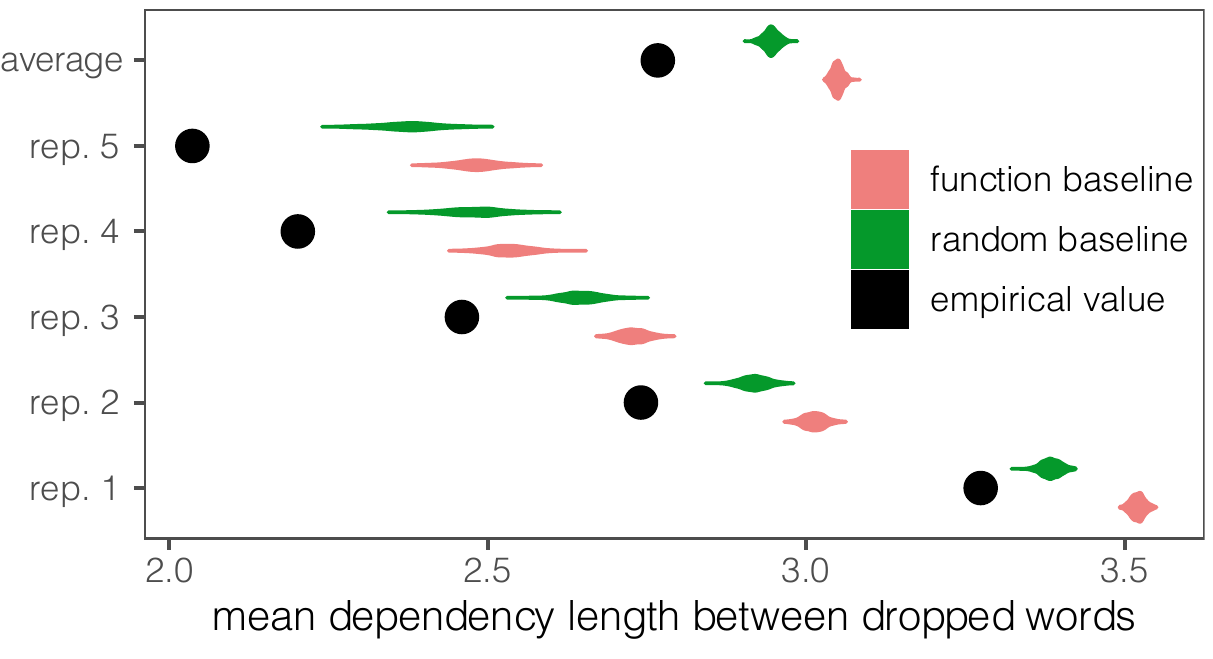}
\caption{The empirical dependency lengths between dropped words are lower than expected under two baselines for every repetition block. Samples from the baselines are shown as densities. }
\label{fig:dependency_length_by_rep}
\end{figure}

\begin{figure}[h]
\centering
\includegraphics[scale=.53]{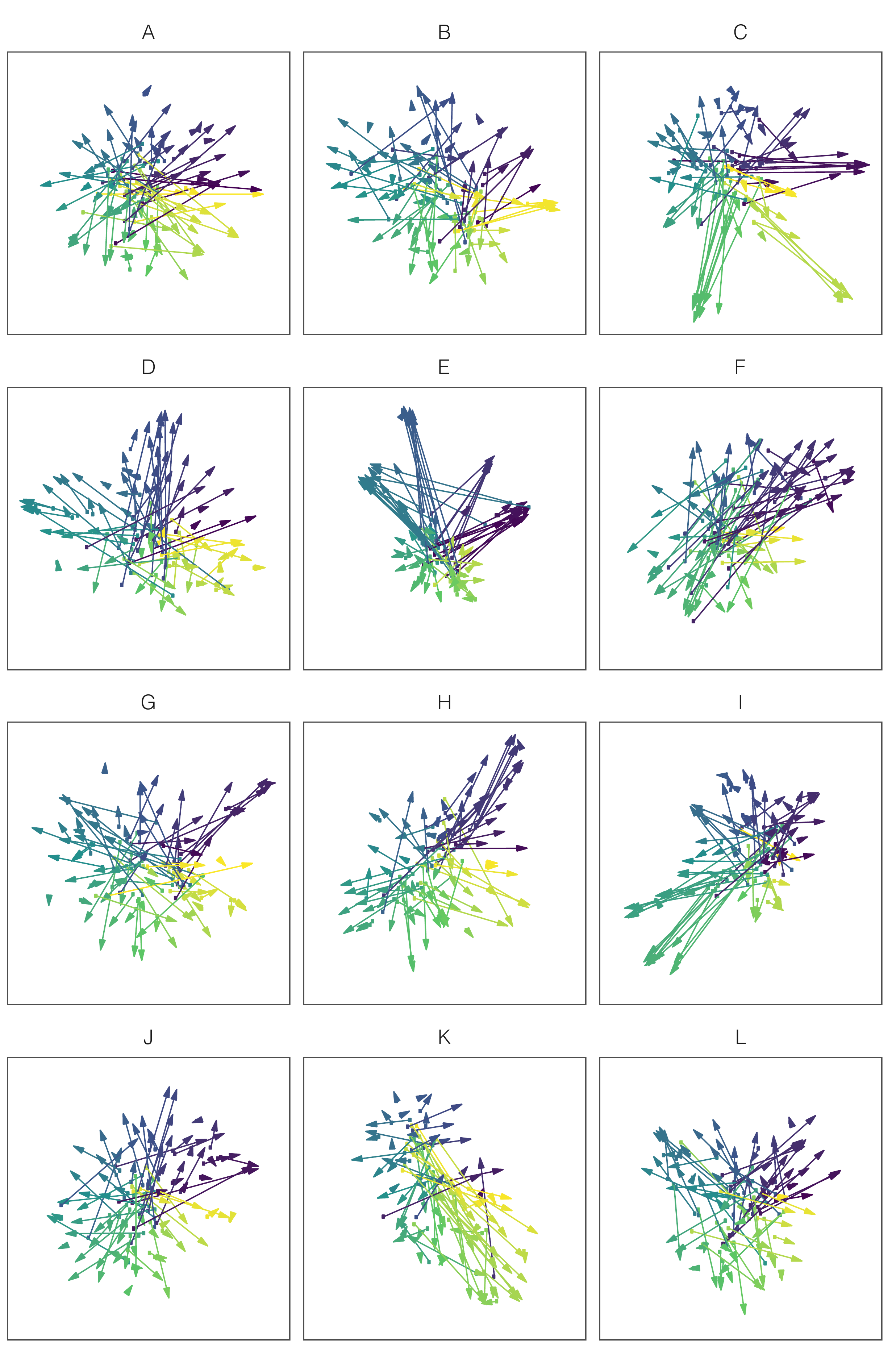}
\caption{t-SNE visualizations of utterance trajectories for all 12 tangrams; panel C is annotated in Fig \ref{fig:tsne}. }
\label{fig:all-tsne}
\end{figure}



\end{document}